\documentclass{ceurart}
\usepackage[utf8]{inputenc}
\usepackage{url}
\usepackage{pdflscape}
\usepackage{rotating}
\usepackage{multirow}
\usepackage{xcolor}

\begin{document}

\copyrightyear{2020}
\copyrightclause{Copyright for this paper by its authors.\\
  Use permitted under Creative Commons License Attribution 4.0
  International (CC BY 4.0).}

\conference{Proceedings of the Iberian Languages Evaluation Forum (IberLEF 2020)}

\title{Overview of CAPITEL Shared Tasks at IberLEF 2020: Named Entity Recognition and Universal Dependencies Parsing}

\author[1]{Jordi Porta-Zamorano}[%
orcid=0000-0001-5620-4916,
]
\ead{porta@rae.es}
\address[1]{Centro de Estudios de la Real Academia Española, Madrid, Spain}

\author[2]{Luis Espinosa-Anke}[%
orcid=0000-0001-6830-9176,
]
\ead{espinosa-anke@cardiff.ac.uk}

\address[2]{School of Computer Science and Informatics, Cardiff University, UK}

\begin{abstract}
We present the results of the CAPITEL-EVAL shared task, held in the context of the IberLEF 2020 competition series. CAPITEL-EVAL consisted on two subtasks: (1) Named Entity Recognition and Classification and (2) Universal Dependency parsing. For both, the source data was a newly annotated corpus, CAPITEL, a collection of Spanish articles in the newswire domain. A total of seven teams participated in CAPITEL-EVAL, with a total of 13 runs submitted across all subtasks. Data, results and further information about this task can be found at \url{sites.google.com/view/capitel2020}.
\end{abstract}


\maketitle

\setcounter{page}{31}
\pagestyle{plain}

\section{Introduction}

Within the framework of the Spanish National Plan for the Advancement of Language Technologies (PlanTL\footnote{\url{https://www.plantl.gob.es}}), the Royal Spanish Academy (RAE) and the Secretariat of State for Digital Advancement (SEAD) of the Ministry of Economy signed an agreement for developing a linguistically annotated corpus of Spanish news articles, aimed at expanding the language resource infrastructure for the Spanish language. The name of this corpus is CAPITEL (\textit{Corpus del Plan de Impulso a las Tecnologías del Lenguaje}), and is composed of contemporary news articles thanks to agreements with a number of news media providers. CAPITEL has three levels of linguistic annotation: morphosyntactic (with lemmas and Universal Dependencies-style POS tags and features), syntactic (following Universal Dependencies v2), and named entities. 

The linguistic annotation of a subset of the CAPITEL corpus has been revised using a machine-annotation-followed-by-human-revision procedure. Manual revision has been carried out by a team of graduated linguists following a set of Annotation Guidelines created specifically for CAPITEL. The named entity and syntactic layers of revised annotations comprise about 1 million words for the former, and roughly 300,000 for the latter.  Due to the size of the corpus and the nature of the annotations, we proposed two IberLEF sub-tasks under the more general, umbrella task of CAPITEL @ IberLEF 2020: 
(1) Named Entity Recognition and Classification and 
(2) Universal Dependency Parsing.


\section{Sub-task 1: NERC}

\subsection{Description}
\label{sect:nerc.sys.desc}

Information extraction tasks, formalized in the late 1980s, are designed to evaluate systems which capture information present in free text, with the goal of enabling better and faster information and content access. One important subset of this information comprises named entities (NE), which, roughly speaking, are textual elements corresponding to names of people, places, organizations and others. Three processes can be applied to NEs: recognition or identification (NER), categorization, i.e., assigning a type according to a predefined set of semantic categories (NERC), and linking, which consists of disambiguating the in-text mention against a knowledge base or sense inventory (NEL).
Since their advent, NER tasks have had notable success, but despite the relative maturity of this subfield, work and research continues to evolve, and new techniques and models appear alongside challenging datasets in different languages, domains and textual genres. The aim of this sub-task, thus, was to challenge participants to apply their systems or solutions to the problem of identifying and classifying NEs in Spanish news articles. This two-stage process falls within the NERC evaluation framework.

The following NE categories were evaluated: Person (PER), Location (LOC), Organization (ORG) and Other (OTH) as defined in the Annotation Guidelines \cite{CAPITEL.NERC} that were shared with participants. The criteria for the identification and classification of entities were based on the capitalization chapter of the Spanish language orthography \cite{OLE}. The contextual meaning has been considered in the classification of entities, so that an entity such as \emph{Madrid} can be classified as PER (a surname), LOC (the city), ORG (the football team) of even OTH (a book title). Moreover, in terms of nesting, only the longest-spanning entities were considered, 
and coordinated entities are considered one single entity except for those where the name indicating the nature of the NE is used in plural to introduce several entities ([\emph{Islas Baleares}]$_{\textsc{loc}}$ \emph{y} [\emph{Canarias}]$_{\textsc{loc}}$).

\subsection{Dataset}

A one-million-word subset of the CAPITEL corpus was randomly sampled into three subsets: training, development and test. The training set comprises 60\% of the corpus, whereas the development and test sets roughly amount to 20\% each. Descriptive statistics for these splits are provided in Table~\ref{tab:nerc.data}. Together with the test set release, an additional collection of documents (background set) was delivered to ensure that participating teams were not be able to perform manual corrections, and also to encourage features such as scalability to larger data collections. Finally, all documents were tokenized and tagged with NEs following an IOBES format.

\begin{table}
\centering
\caption{Description of the data for CAPITEL sub-task 1: NERC}
\label{tab:nerc.data}
\begin{tabular}{|c|rrrr|r|r|}\hline
\multicolumn{1}{|c|}{\textbf{Dataset}}
& \multicolumn{1}{c}{\textbf{PER}}
& \multicolumn{1}{c}{\textbf{LOC}}
& \multicolumn{1}{c}{\textbf{ORG}}
& \multicolumn{1}{c}{\textbf{OTH}}
& \multicolumn{1}{|c|}{\textbf{Sents.}}
& \multicolumn{1}{c|}{\textbf{Tokens}}
\\\hline
train & 9,087 & 7,513 & 9,285 & 5,426 & 22,647 & 606,418 \\
devel & 2,900 & 2,490 & 3,058 & 1,781 &  7,549 & 202,408 \\
test  & 2,996 & 2,348 & 3,143 & 1,739 &  7,549 & 199,773 \\
\hline
\textbf{Total} & 14,983 & 12,351 & 15,486 & 8,946 & 37,745 & 1,008,599 \\
\hline
\end{tabular}
\end{table}

\subsection{Evaluation Metrics}

The metrics used for evaluation were Precision (the percentage of named entities in the system's output that are correctly recognized and classified), Recall (the percentage of named entities in the test set that were correctly recognized and classified) and macro averaged F${_1}$ score (the harmonic mean of Precision and Recall), with the latter being used as the official evaluation score and for the final ranking of the participating teams.

\subsection{Systems and Results}

We had 22 registrations, 5 final participants with 9 systems submitted and 4 system descriptions.

The Ragerri Team from HiTZ Center-Ixa UPV/EHU presents in \cite{capitel2020-2} the combination of several systems based on Flair \cite{akbik-etal-2018-contextual} and Transformer architectures \cite{NIPS2017_7181}. They perform experiments with Multilingual BERT (mBERT), XML-RoBERTa (base), BETO (BETO is a BERT-based model pre-trained with Spanish texts \cite{CaneteCFP2020}), and Flair off-the-shelf models for Spanish and a monolingual model trained with the OSCAR corpus. All the individual systems' F$_1$ were within 88.29-89.95\% and the combination of five of them using a simple agreement scheme of three achieved the first rank with a 90.30\% F$_1$. 

The Vicomtech Team presents in \cite{capitel2020-3} a system based on the BERT architecture and several experiments using multilingual BERT (mBERT) and BETO pre-trained models. BERT models are used to give each token a contextual embedding that are then passed to a fully connected layer to classify each of these tokens. Their work addresses also several interesting issues with the BETO vocabulary and tokenizer, namely: punctuation marks missing in the tokenizer's vocabulary and problems with certain diacritics and characters. Their systems were fine-tuned with CAPITEL training data and results were 2-3\% F${_1}$ lower than the best performing system.

The Yanghao Team from Huawei Translation Service Center presents in \cite{capitel2020-1} a system that uses Multilingual BERT as encoder and a linear layer as a classifier, and is trained with additional 38,000 sentences from WMT news translation corpus \cite{TIEDEMANN12.463} annotated using Spacy \cite{spacy2}. Their experimental results suggest that pre-training with the augmented set and then fine-tune on CAPITEL improves performance when compared to training on any of them separately or mixed. 


The Lirondos Team from ISI-USC presents in \cite{capitel2020-5} two sequence labelling systems: A CRF model with handcrafted features and a BiLSTM-CRF model with word and character embeddings. A feature ablation study demonstrated that all features contribute positively to the CRF model, with word embeddings being the most informative feature, yielding an 
F{$_1$} score of 84.39\%. On the other hand, their 
BiLSTM-CRF model obtained an F{$_1$} score of 83.01\%. A interesting error analysis has shown that many of the errors correspond to OTH entities, contextual annotation of some entities (OTH versus ORG or LOC versus ORG), nested entities, and person nicknames with unusual typographical shapes.

Finally, the LolaZarra Team was ranked the last and did not submit any system description paper.

\begin{table*}
\centering
\caption{Results of the CAPITEL sub-task 1: NERC.}\label{tab:nerc}
\begin{tabular}{|c|c|c|c|cccc|c|c|}\hline
\multicolumn{1}{|c|}{\textbf{Rank}}&
\multicolumn{1}{c|}{\textbf{Team}}&
\multicolumn{1}{|c|}{\textbf{Ref.}}&
\multicolumn{1}{c|}{\textbf{Metric}}&
\multicolumn{1}{c}{\textbf{PER}} &
\multicolumn{1}{c}{\textbf{LOC}} &   
\multicolumn{1}{c}{\textbf{ORG}} &    
\multicolumn{1}{c|}{\textbf{OTH}} &    
\multicolumn{1}{c|}{\textbf{Micro}} &     
\multicolumn{1}{c|}{\textbf{Macro}} 
\\\hline

\multirow{3}{*}{(1)} & \multirow{3}{*}{ragerri} & \multirow{3}{*}{\cite{capitel2020-2}} &
P        & 96.40 & 90.47 & 88.63 & 83.36 & 90.50 & 90.43 \\
&&&R     & 97.46 & 91.74 & 87.31 & 80.68 & 90.17 & 90.17 \\
&&&F$_1$ & 96.93 & 91.10 & 87.96 & 82.00 & 90.34 & 90.30 \\
\hline

\multirow{3}{*}{(2)} & \multirow{3}{*}{ragerri} & \multirow{3}{*}{\cite{capitel2020-2}} &
P        & 96.50 & 90.19  & 88.05 & 84.37 & 90.46 & 90.39 \\
&&&R     & 97.46 & 91.27  & 87.21 & 81.02 & 90.09 & 90.09 \\
&&&F$_1$ & 96.98 & 90.73  & 87.63 & 82.66 & 90.27 & 90.23 \\
\hline

\multirow{3}{*}{(3)} & \multirow{3}{*}{ragerri} & \multirow{3}{*}{\cite{capitel2020-2}} &
P        & 96.69 & 90.56 & 88.03 & 83.39 & 90.42 & 90.36 \\
&&&R     & 97.60 & 91.14 & 87.24 & 80.56 & 90.04 & 90.04 \\
&&&F$_1$ & 97.14 & 90.85 & 87.63 & 81.95 & 90.23 & 90.19 \\ 
\hline

\multirow{3}{*}{(4)} & \multirow{3}{*}{mcuadros} &  \multirow{3}{*}{\cite{capitel2020-3}} & 
P        & 93.48 & 89.36 & 85.76 & 79.63 & 87.88 & 87.81 \\
&&&R     & 96.70 & 88.03 & 85.87 & 77.34 & 88.09 & 88.09 \\
&&&F$_1$ & 95.06 & 88.69 & 85.82 & 78.47 & 87.99 & 87.94 \\
\hline

\multirow{3}{*}{(5)} & \multirow{3}{*}{yanghao} & \multirow{3}{*}{\cite{capitel2020-1}} & 
P        & 94.30 & 87.30 & 84.99 & 79.52 & 87.38 & 87.32 \\
&&&R     & 96.16 & 89.86 & 85.94 & 77.69 & 88.43 & 88.43 \\
&&&F$_1$ & 95.22 & 88.56 & 85.46 & 78.59 & 87.90 & 87.87 \\ 
\hline

\multirow{3}{*}{(6)} & \multirow{3}{*}{lirondos} & \multirow{3}{*}{\cite{capitel2020-5}} &
P        & 92.48 & 83.42 & 83.76 & 75.03 & 84.93 & 84.75 \\
&&&R     & 94.46 & 86.97 & 80.43 & 69.12 & 84.12 & 84.12 \\ 
&&&F$_1$ & 93.46 & 85.15 & 82.06 & 71.95 & 84.52 & 84.39 \\
\hline

\multirow{3}{*}{(7)} & \multirow{3}{*}{LolaZarra} & \multirow{3}{*}{-} &
P        & 91.52 & 83.39 & 80.10 & 78.31 & 83.93 & 83.90 \\
&&&R     & 92.62 & 80.41 & 83.39 & 73.72 & 83.77 & 83.77 \\ 
&&&F$_1$ & 92.07 & 81.87 & 81.71 & 75.95 & 83.85 & 83.80 \\
\hline

\multirow{3}{*}{(8)} & \multirow{3}{*}{lirondos} & \multirow{3}{*}{\cite{capitel2020-5}} &
P        & 94.37 & 85.68 & 84.20 & 65.47 & 83.93 & 84.33 \\ 
&&&R     & 90.72 & 83.35 & 78.14 & 71.08 & 81.82 & 81.82 \\
&&&F$_1$ & 92.51 & 84.50 & 81.06 & 68.16 & 82.86 & 83.01 \\
\hline

\multirow{3}{*}{(9)} & \multirow{3}{*}{lirondos} & \multirow{3}{*}{\cite{capitel2020-5}} &
P        & 93.23 & 82.05 & 84.55 & 63.89 & 82.67 & 83.01 \\
&&&R     & 90.09 & 82.54 & 73.85 & 67.17 & 79.46 & 79.46 \\
&&&F$_1$ & 91.63 & 82.29 & 78.84 & 65.49 & 81.03 & 81.11 \\
\hline
\end{tabular}
\end{table*}

\section{Sub-task 2: UD Parsing}

\subsection{Description}

Dependency-based syntactic parsing has become popular in NLP in recent years. One of the reasons for this popularity is the transparent encoding of predicate-argument structures, which is useful in many downstream applications. Another reason is that it is better suited than phrase-structure grammars for languages with free or flexible word order.
Universal Dependencies (UD) is a framework for consistent annotation of grammar (parts of speech, morphological features and syntactic dependencies) across different human languages. Moreover, the UD initiative is an open community effort with over 200 contributors which has produced more than 100 treebanks in over 70 languages. 

The aim of this sub-task was to challenge participants to apply their systems or solutions to the problem of Universal Dependency parsing of Spanish news articles as defined in the Annotation Guidelines for the CAPITEL corpus \cite{CAPITEL.UD}.

\subsection{Dataset}

A 300,000-word subset of CAPITEL was provided for this sub-task. In addition to head and dependency relations in CoNLL-U format, this subset was also tokenized and annotated with lemmas and UD tags and features. 
Similarly to the NERC dataset, we randomly sampled it into three subsets: training, development and test. The training set comprises about 50\% of the corpus, whereas the development and test sets roughly amount to 25\% each. The description of the data sets can be found in Table~\ref{tab:ud.data}. In addition, the distribution of labels in the test set is given in Table~\ref{tab:ud.details} along with the results of the sub-task. Together with the test set release, an additional collection of documents (background set) were included to ensure that participating teams were not be able to perform manual corrections, and also to encourage features such as scalability to larger data collections.

\begin{table}
\centering
\caption{Description of the data for CAPITEL sub-task 2: UD Parsing}
\label{tab:ud.data}
\begin{tabular}{|c|r|r|}\hline
\multicolumn{1}{|c}{\textbf{Dataset}}
& \multicolumn{1}{|c|}{\textbf{Sents.}}
& \multicolumn{1}{c|}{\textbf{Tokens}}\\
\hline
train &  7,086 & 185,560 \\
devel &  2,362 &  61,137 \\
test  &  2,363 &  62,682 \\
\hline
\textbf{Total} & 11,811 & 309,379 \\
\hline
\end{tabular}
\end{table}

\subsection{Evaluation Metrics}

The metrics for the evaluation phase were Unlabeled Attachment Score (UAS): The percentage of words that have the correct head, and Labeled Attachment Score (LAS): The percentage of words that have the correct head and dependency label, with the latter being used as the official evaluation score, and for the final ranking of the participating teams.

\subsection{Systems and Results}

We had, in this subtask, 12 registrations, 2 final participants with 4 submitted systems and 2 system descriptions.

The Vicomtech Team presents in \cite{capitel2020-3} a system based on the BERT architecture and several experiments using multilingual BERT (mBERT) and BETO pre-trained models.
BERT models are used to encode a matrix of all-vs-all token encoding vectors and then pass to several classification layers predicting the connectivity of tokens and their relation types. Their work addresses also some issues that had been explained in \ref{sect:nerc.sys.desc}. Their systems were fine-tuned with CAPITEL training data and results on the development set were slightly better using BETO (UAS: 91.540, LAS: 88.410) instead of mBERT (UAS: 91.220, LAS: 87.860), so 
only the BETO results were submitted as their official run.

MartínLendinez Team presents in \cite{capitel2020-4} the combination of the output of different UD parsing toolkits using a voting scheme and the augmentation of the training set with 14,305 annotated sentences from the AnCora annotated corpus \cite{taule-etal-2008-ancora}.\footnote{There is also a discussion on some differences in terms of tokenization and analysis between 
CAPITEL and AnCora.} Three different toolkits were selected not because of their performance in similar tasks but for their accessibility and documentation. 
These toolkits were UDPipe \cite{udpipe:2017}, NLP-Cube \cite{boros-etal-2018-nlp} and Stanza \cite{qi2020stanza}. 
As we can see in the summary provided in Table~\ref{tab:ud}, the final submitted results were obtained with Stanza trained on CAPITEL (4), Stanza trained on CAPITEL and AnCora (3), and the combination of the previous two plus NLP-Cube trained on CAPITEL (1).

\begin{table}
\caption{Results of the CAPITEL sub-task 2: UD Parsing}\label{tab:ud}
\centering
\begin{tabular}{|c|c|c|c|c|}\hline
\multicolumn{1}{|c|}{\textbf{Rank}} &
\multicolumn{1}{c|}{\textbf{Team}} &
\multicolumn{1}{|c|}{\textbf{Ref.}} &
\multicolumn{1}{c|}{\textbf{UAS}} &
\multicolumn{1}{c|}{\textbf{LAS}}
\\\hline
(1) & MartinLendinez (CACV) & \cite{capitel2020-4} & 91.935 & 88.660 \\
(2) & Vicomtech (BETO) & \cite{capitel2020-3} & 91.875 & 88.600 \\
(3) & MartinLendinez (CA) & \cite{capitel2020-4} & 91.773 & 88.531 \\
(4) & MartinLendinez (C) & \cite{capitel2020-4} & 91.715 & 88.467 \\
\hline
\end{tabular}
\end{table}

As it can be seen in Table~\ref{tab:ud}, results on this sub-task are very tight, with first and second systems being only 0.06\% apart, and with only 0.193\% between first and fourth. 
The submission by MartínLendinez was the highest ranked, and Vicomtech the simplest, and acknowledged and described by the authors as a sort of BERT-based baseline. We provide a breakdown of the results by relation type in Table~\ref{tab:ud.details}.

\begin{table*}
\caption{Detailed results of the CAPITEL sub-task 2: UD Parsing}\label{tab:ud.details}
\centering
\begin{tabular}{|c|r|rr|rr|rr|rr|}\cline{3-10}
\multicolumn{2}{c}{} &
\multicolumn{2}{|c|}{\textbf{(1)}} &
\multicolumn{2}{c|}{\textbf{(2)}} &
\multicolumn{2}{c|}{\textbf{(3)}} &
\multicolumn{2}{c|}{\textbf{(4)}} 
\\\hline
\multicolumn{1}{|c|}{\textbf{Label}} &
\multicolumn{1}{c|}{\textbf{Freq.}} &
\multicolumn{1}{c}{\textbf{UAS}} &
\multicolumn{1}{c|}{\textbf{LAS}} &
\multicolumn{1}{c}{\textbf{UAS}} &
\multicolumn{1}{c|}{\textbf{LAS}} &
\multicolumn{1}{c}{\textbf{UAS}} &
\multicolumn{1}{c|}{\textbf{LAS}} &
\multicolumn{1}{c}{\textbf{UAS}} &
\multicolumn{1}{c|}{\textbf{LAS}} \\
\hline
acl        &   501 & 80.04 & 65.47 & 82.44 & 67.27 & 80.04 & 65.67 & 79.04 & 65.27 \\
acl:relcl  & 1,004 & 74.90 & 73.61 & 79.68 & 78.19 & 75.50 & 74.30 & 74.70 & 73.41\\
advcl      & 1,004 & 79.18 & 73.11 & 78.29 & 71.02 & 78.69 & 73.11 & 79.18 & 73.61\\
advmod     & 2,062 & 87.92 & 85.01 & 87.34 & 83.75 & 86.23 & 83.51  & 87.20 & 83.95\\ 
amod       & 3,228 & 96.96 & 94.95 & 96.78 & 94.24 & 96.81 & 94.76 & 97.00 & 95.26 \\
appos      & 1,090 & 85.50 & 77.61 & 84.22 & 74.50 & 84.40 & 76.70  & 84.86 & 76.79\\
aux        &   259 & 40.93 & 31.27 & 52.51 & 46.72 & 48.26 & 38.22 & 40.93 & 30.89 \\
aux:pass   &    56 & 100.00 & 94.64 & 98.21 & 83.93 & 100.00 & 94.64 & 100.00 & 94.64 \\
case       & 8,705 & 98.99 & 98.70 & 98.78 & 98.24 & 98.83 & 98.55 & 99.05 & 98.74 \\
cc         & 2,018 & 95.29 & 92.86 & 95.00 & 92.72 & 95.14 & 92.77 & 94.50 & 92.02 \\
ccomp      &   399 & 90.73 & 83.71 & 90.48 & 84.21 & 90.73 & 83.96 & 88.72 & 81.45 \\
compound   &    22 & 63.64 & 18.18 & 81.82 & 45.45 & 59.09 & 13.64 & 68.18 & 27.27 \\
conj       & 2,361 & 76.20 & 73.27 & 77.42 & 74.29 & 76.37 & 73.36 & 75.60 & 72.47 \\
cop        &   925 & 93.30 & 89.84 & 93.19 & 89.84 & 93.95 & 90.49 & 92.76 & 89.51 \\
csubj      &   111 & 81.08 & 60.36 & 83.78 & 63.96 & 82.88 & 62.16 & 79.28 & 52.25 \\
dep        &    28 & 75.00 & 7.14 & 67.86 & 3.57 & 75.00 & 7.14 & 71.43 & 7.14 \\
det        & 8,840 & 99.42 & 99.37 & 99.29 & 99.17 & 99.33 & 99.29 & 99.38 & 99.32 \\
discourse  &    36 & 80.56 & 2.78 & 86.11 & 8.33 & 77.78 & 2.78 & 77.78 & 5.56 \\
expl       &    46 & 97.83 & 6.52 & 95.65 & 41.30 & 97.83 & 6.52 & 97.83 & 23.91 \\
expl:impers & 29 & 93.10 & 6.90 & 86.21 & 20.69 & 93.10 & 6.90 & 89.66 & 10.34 \\
expl:pass  &  360 & 99.44 & 75.56 & 99.17 & 82.50 & 99.44 & 75.56 & 98.89 & 79.72 \\
expl:pv    &  343 & 97.67 & 70.85 & 97.67 & 74.05 & 97.38 & 70.55 & 97.38 & 68.22 \\
fixed      &  219 & 71.23 & 68.04 & 71.23 & 65.75 & 70.32 & 66.67 & 71.23 & 68.49 \\
flat       & 130 & 92.31 & 45.38 & 88.46 & 53.85 & 91.54 & 45.38 & 90.77 & 50.00 \\
flat:foreign & 409 & 90.71 & 86.80 & 78.48 & 70.17 & 91.44 & 87.53 & 91.69 & 87.29 \\
goeswith   & 2 & 0.00 & 0.00 & 0.00 & 0.00 & 0.00 & 0.00 & 0.00 & 0.00 \\
iobj       & 329 & 94.22 & 77.51 & 91.19 & 72.95 & 93.62 & 76.90 & 92.40 & 69.91 \\
mark       & 1,992 & 91.62 & 85.99 & 92.67 & 86.75 & 92.12 & 86.40 & 91.47 & 85.94 \\
mark:iobj  & 9 & 100.00 & 33.33 & 88.89 & 44.44 & 100.00 & 33.33 & 100.00 & 22.22 \\
mark:mod   & 282 & 93.97 & 79.79 & 93.97 & 83.69 & 93.62 & 79.43 & 94.33 & 80.85 \\
mark:obj   & 119 & 89.92 & 49.58 & 89.92 & 55.46 & 90.76 & 50.42 & 90.76 & 42.86 \\
mark:subj  & 816 & 94.24 & 87.99 & 95.10 & 87.01 & 94.61 & 88.36 & 93.87 & 88.36 \\
nmod       & 4,609 & 88.24 & 87.18 & 89.48 & 88.15 & 88.28 & 87.22 & 87.85 & 86.77 \\
nsubj      & 2,302 & 93.61 & 89.27 & 93.53 & 89.27 & 93.61 & 89.23 & 93.01 & 86.92 \\
nsubj:pass & 29 & 96.55 & 58.62 & 96.55 & 55.17 & 96.55 & 58.62 & 96.55 & 55.17 \\
nummod & 689 & 97.68 & 96.66 & 97.68 & 95.36 & 97.53 & 96.37 & 97.97 & 96.37 \\
obj & 2,235  & 98.61 & 89.80 & 97.67 & 90.65 & 98.39 & 89.75 & 98.30 & 91.50 \\
obl & 3,298  & 87.54 & 81.96 & 87.17 & 81.78 & 87.48 & 82.05 & 87.72 & 82.20 \\
obl:agent    & 94 & 98.94 & 86.17 & 97.87 & 85.11 & 100.00 & 87.23 & 97.87 & 90.43 \\
orphan       & 17 & 70.59 & 0.00 & 70.59 & 0.00 & 70.59 & 0.00 & 64.71 & 0.00 \\
parataxis    & 881 & 74.01 & 59.82 & 72.76 & 60.39 & 72.99 & 58.80 & 74.35 & 61.29 \\
punct        & 8,028 & 89.09 & 88.95 & 88.17 & 88.02 & 88.64 & 88.52 & 88.78 & 88.61 \\
root         & 2,394 & 93.27 & 93.07 & 93.48 & 93.32 & 93.23 & 93.02 & 92.86 & 92.65 \\
xcomp        & 372 & 75.81 & 69.09 & 80.38 & 72.31 & 72.04 & 65.59 & 79.03 & 71.51 \\
\hline
\multicolumn{1}{|c|}{\textbf{Total}} & 
62,682 & 91.935 & 88.660 & 91.875 & 88.600 & 91.773 & 88.531 & 91.715 & 88.467 \\
\hline
\end{tabular}
\end{table*}

\section{Conclusions}
 
 Most of the submitted systems obtained good results overall. In both sub-tasks, the majority of them uses BERT, either multilingual or monolingual and some systems combines the output of several models. Also the augmentation of data from other corpora, or produced by other annotation systems, added to the training data or used to fine-tune the models, despite the heterogeneity of the annotations or domain differences have shown some modest improvements.
 
\section{Acknowledgements}

We would like to thank specially to 
David Pérez Fernández, Doaa Samy, and all the people involved in the PlanTL, for their contribution in making these shared tasks possible, and José-Luis Sancho-Sánchez and Rafael-J. Ureña-Ruiz from the Centro de Estudios de la RAE for their help in preparing the data. We would also like to thank the task participants who provided helpful inputs to improve the quality of the dataset and the task itself.

\bibliography{capitel2020}

\end{document}